\documentclass{article} % For LaTeX2e
\usepackage{iclr2027_conference,times}

\usepackage{amsmath,amsfonts,bm}

\def\eqref#1{equation~\ref{#1}}
\def\1{\bm{1}}

\DeclareMathAlphabet{\mathsfit}{\encodingdefault}{\sfdefault}{m}{sl}
\SetMathAlphabet{\mathsfit}{bold}{\encodingdefault}{\sfdefault}{bx}{n}

\usepackage{hyperref}
\usepackage{url}
\usepackage{graphicx}
\usepackage{booktabs}
\graphicspath{{pics/}}

\title{InstructTTS-OPD: On-Policy Distillation for Multi-Objective Instruction-Following TTS}

\author{%
Haitao Li\textsuperscript{1,2}, Tian Tan\textsuperscript{3},
Yuguang Yang\textsuperscript{5}, Guanrou Yang\textsuperscript{3,2},\\
Wenhao Guan\textsuperscript{4,2}, Chenglin Li\textsuperscript{1,2},
Shan Yang\textsuperscript{5}, Xie Chen\textsuperscript{3,2}\\[0.5ex]
\normalfont\textsuperscript{1}Zhejiang University\quad
\textsuperscript{2}Shanghai Innovation Institute\\
\normalfont\textsuperscript{3}Shanghai Jiao Tong University\quad
\textsuperscript{4}Xiamen University\quad
\textsuperscript{5}Tencent Hunyuan
}

\newcommand{\methodname}{InstructTTS-OPD}

\iclrfinalcopy
\begin{document}

\maketitle

\begin{abstract}
Instruction-following text-to-speech (Instruct-TTS) renders a target transcript in the manner described by a free-form natural-language instruction, and a good rendition must succeed on three fronts at once: following the instruction, pronouncing the transcript correctly, and sounding natural. Reinforcement learning has become the standard way to post-train such generative models toward human preferences, and TTS systems increasingly adopt it. When the three objectives are optimized together under a single scalar reward, typically with group relative policy optimization (GRPO), training tends to be unstable: the objectives compete, and the shared update drifts toward whichever one is easiest to improve, producing a seesaw among them. We instead adapt on-policy distillation (OPD), which fuses several single-objective teachers by matching their velocity fields on the student's own rollouts. Realizing this for Instruct-TTS first calls for a sound reward on each axis, which we analyze carefully: a distilled Qwen-Omni proxy for the costly Gemini judge, dialect-free data for the WER reward, and DNSMOS in place of the language-biased UTMOS. We then train three single-objective teachers with GRPO and fuse them with OPD, routing each rollout to the instruction-following or WER teacher by its data domain and letting every rollout additionally follow the MOS teacher for quality. This lets instruction following and intelligibility improve together without a trade-off, but the quality target conflicts with instruction following, so following it additively reintroduces the seesaw. We therefore propose conflict-aware velocity fusion, which projects the quality target off the direction that opposes the primary objective before fusing, and show that it markedly reduces the instruction-following/quality trade-off. The fused model improves over scalar-reward GRPO on all three objectives at once. Demo is available at \url{https://cucl-2.github.io/InstructTTS-OPD}.
% Code is available at \url{https://anonymous.4open.science/r/InstructTTS-OPD}.
\end{abstract}

\section{Introduction}

Text-to-speech (TTS) has advanced to the point where the synthesized voice is often hard to tell from a real recording~\citep{anastassiou2024seed, du2025cosyvoice, hu2026qwen3, xu2026x, liu2026cross}, and attention is shifting from merely reading a transcript to controlling \emph{how} it is read~\citep{li2026restyle, yang2024instructtts}. Instruction-following TTS (Instruct-TTS) captures this goal: given a target transcript together with a free-form natural-language instruction, such as ``\emph{read this in a trembling, sobbing voice}'' or ``\emph{speak the following line in Cantonese, sounding impatient},'' the model must render the transcript in the manner described~\citep{guo2023prompttts, yang2024instructtts}. A good rendition has to succeed on three fronts at once: it must follow the instruction (instruction following, IF), pronounce the transcript correctly (intelligibility, measured by word or character error rate, WER/CER), and sound natural (quality, measured by mean opinion score, MOS). Among the three, instruction following is primary, as it distinguishes Instruct-TTS from ordinary TTS.

TTS models are increasingly post-trained with reinforcement learning (RL) to align their outputs with human preference~\citep{ma2026auk, zhang2024speechalign, wang2026flowtts, yang2026grow}, and the dominant recipe is group relative policy optimization (GRPO)~\citep{shao2024deepseekmath, liu2026flow} with a scalar reward. When several objectives must be optimized together, they are collapsed into a single number, and this scalarization discards directional information: when two objectives disagree on how the output should change, their contributions partially cancel, and a single scalar cannot express which trade-off to make. Optimization then drifts toward whichever objective is easiest to improve, so a softer objective comes to dominate the shared update while the competing ones regress. The result is the familiar seesaw among the three objectives, and reweighting the scalar only shifts the balance point rather than removing the tension.

A promising alternative comes from image generation, where on-policy distillation (OPD) fuses several single-objective expert teachers by matching their velocity fields on the student's own rollouts~\citep{fang2026flow}, giving a dense, per-step supervision signal in place of a single scalar. Bringing this idea to Instruct-TTS is not immediate, because the task carries characteristics that the image setting does not, and it first calls for a careful choice of reward along each of the three axes. For instruction following, the reliable judge is Gemini~\citep{comanici2025gemini}, but querying it during training is prohibitively expensive; we therefore distill a Qwen-Omni~\citep{xu2025qwen25omni} reward model that closely tracks the Gemini judgment and serves as a low-cost replacement. For intelligibility, we find that the WER teacher must be trained on dialect-free data: otherwise the model learns to lower WER by not speaking the requested dialect, which in turn drags down IF. For quality, we show that UTMOS~\citep{saeki2022utmos} is a poor choice because it carries a strong bias between Chinese and English, which pushes the model toward a foreign-accented reading of Chinese, and we adopt DNSMOS~\citep{reddy2022dnsmos} instead.

With these rewards in place, we train three single-objective teachers with GRPO and fuse them through OPD. We route each rollout to the IF or WER teacher by its data domain, and let every rollout additionally follow the MOS teacher for quality. Routing the two primary objectives to disjoint utterances lets instruction following and intelligibility improve together without a trade-off. Quality, however, is applied to every rollout and, as we show, conflicts with instruction following, so following it additively reintroduces the seesaw. We therefore propose conflict-aware velocity fusion, which projects the quality target off the direction that opposes the primary objective before fusing, a per-frame geometric operation on the distillation signal that a scalar reward cannot express. This markedly reduces the instruction-following/quality trade-off, and the fused model improves over scalar-reward GRPO on all three axes at once.

The main contributions of this work are summarized as follows:
\begin{itemize}
\item \textbf{Reward analysis for Instruct-TTS.} We identify a sound reward along each axis: a distilled Qwen-Omni proxy for the costly Gemini judge, a dialect-free WER reward, and DNSMOS in place of the language-biased UTMOS.
\item \textbf{On-policy distillation for multi-objective Instruct-TTS.} We fuse three single-objective GRPO teachers with OPD, routing each rollout to one teacher by its data domain, and surpass scalar-reward GRPO on all three axes.
\item \textbf{Conflict-aware velocity fusion.} Because the quality target conflicts with instruction following, we project it off the opposing direction before fusing, a per-frame geometric operation that scalar-reward RL cannot express and that yields the best overall system.
\end{itemize}

\section{Related Work}
\label{sec:related}

\paragraph{Instruction-following TTS.}
Instruct-TTS asks a model to speak a target transcript in the manner described by a free-form natural-language instruction, such as a desired emotion, timbre, speaking rate, or dialect, and a good rendition must follow the instruction, stay intelligible, and sound natural at once. Early systems encode the style description with a dedicated prompt encoder and learn a mapping from the described attributes to acoustic style, as in PromptTTS~\citep{guo2023prompttts} and InstructTTS~\citep{yang2024instructtts}. As large language models improved the understanding of free-form instructions, more recent systems such as FlexiVoice~\citep{chen2026flexivoice}, Mimo-Audio~\citep{xiaomi2025mimo}, MOSS-VoiceGenerator~\citep{huang2026moss}, and Qwen3-TTS~\citep{hu2026qwen3} pair an LLM backbone with a speech decoder and follow much richer, open-vocabulary instructions.

\paragraph{Reinforcement learning for TTS.}
Reinforcement learning is increasingly used to post-train TTS models toward objectives that are not directly differentiable. Early efforts focus on autoregressive, LLM-based TTS in which speech is represented as discrete tokens, so the RL machinery closely follows that of text LLMs. Seed-TTS~\citep{anastassiou2024seed} optimizes WER and speaker-similarity rewards with PPO, REINFORCE, and DPO; SpeechAlign~\citep{zhang2024speechalign} aligns the synthetic distribution with a gold reference via DPO; DiffRO~\citep{gao2025differentiable} adopts a token-level differentiable ASR reward; RRPO~\citep{wang2026rrpo} designs a more robust reward that alleviates reward hacking; and GRPO-TTS~\citep{liu2026group} combines GRPO with a composite CER and NLL reward. A second line performs RL on flow-matching TTS: F5R-TTS~\citep{sun2025f5r} reformulates the deterministic flow into a probabilistic one to make policy gradients available, at the cost of retraining, whereas FlowTTS-GRPO~\citep{wang2026flowtts} turns the sampling ODE into an SDE and fine-tunes an open-source flow model directly with GRPO. Built on a DiTAR-style architecture, our work follows this second line and carries out RL and distillation in the continuous flow-matching space rather than over discrete tokens.

\paragraph{On-policy distillation.}
On-policy distillation (OPD) was first introduced for LLMs, where a student learns to match its teachers on the rollouts it generates itself, and has proven effective at combining several teachers into one student while easing the trade-off among them~\citep{agarwal2024policy, gu2023minillm, ma2026mopd, xu2026deepseek}. The idea was later carried to image generation by Flow-OPD~\citep{fang2026flow} and DiffusionOPD~\citep{li2026diffusionopd}, which recast distillation on a flow-matching or diffusion model as matching the teacher velocity fields along the student's own rollouts; Flow-OPD, for instance, jointly optimizes GenEval, OCR, and aesthetic quality, and Qwen-Image-2.0-RL~\citep{xu2026qwen} applies OPD to merge a generation and an editing model into one. We bring OPD to Instruct-TTS, optimizing instruction following, intelligibility, and quality together; the characteristics of speech introduce challenges the image setting does not face.

\section{Method}
\label{sec:method}

\begin{figure}[t]
\centering
\includegraphics[width=\linewidth]{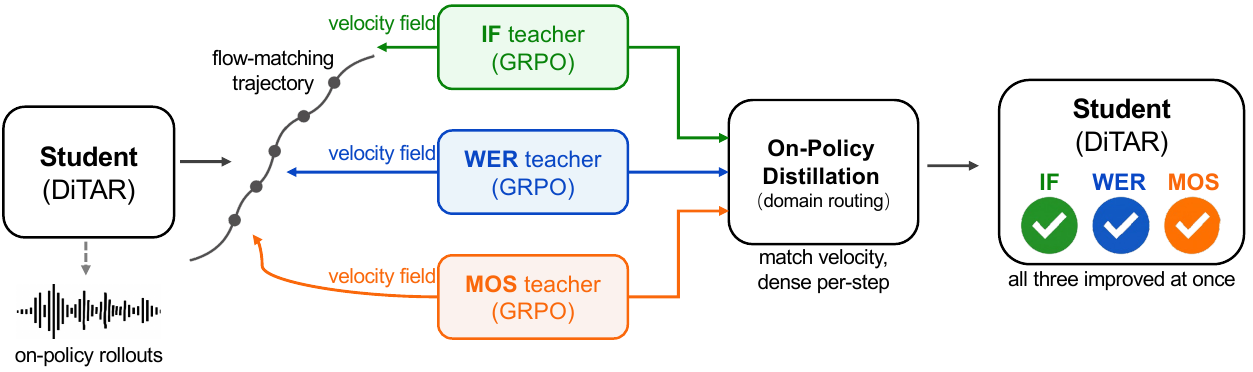}
\caption{Overview of \methodname{}. Three single-objective teachers (IF, WER,
MOS), each trained with GRPO, are evaluated on the student's own on-policy
flow-matching rollouts and supply dense per-step velocity-field targets. Domain
routing and conflict-aware fusion combine these targets so the student improves
all three objectives at once, avoiding the scalar-reward seesaw.}
\label{fig:overview}
\end{figure}

\subsection{Preliminaries: DiTAR backbone and flow matching}
\label{sec:method:prelim}

We build on a DiTAR-style architecture~\citep{jia2025ditar}, which couples an autoregressive language model with a local flow-matching decoder. At each step the language model reads the text together with a summary of the audio generated so far and emits a condition vector $c$; a flow-matching decoder, a DiT block we refer to as the local decoder, turns Gaussian noise into the audio latents of the current frame under that condition; an aggregation encoder compresses the new frame back into a single vector that is fed to the language model as the next input, closing the autoregressive loop. A frozen vocoder finally maps the latents to a waveform. The language model is a $0.5$B Qwen~\citep{yang2025qwen3}; together with the local decoder ($193$M) and the aggregation encoder ($48$M) the DiTAR model totals about $0.7$B parameters (the vocoder is frozen). This design keeps generation autoregressive at the frame level while making each frame the output of a continuous flow-matching process, which is exactly where we apply reinforcement learning and distillation.

Flow matching~\citep{lipman2022flow, liu2022flow} learns a velocity field $v_\theta(x_t, t)$ that transports a Gaussian prior at $t{=}0$ to the data distribution at $t{=}1$. Given noise $x_0$ and a target latent $x_1$, one defines the straight-line path $x_t = (1-t)\,x_0 + t\,x_1$ with corresponding velocity $u_t = x_1 - x_0$, and trains $v_\theta$ to regress $u_t$. At inference, generation starts from noise and integrates the ordinary differential equation (ODE) $\mathrm{d}x = v_\theta(x_t, t)\,\mathrm{d}t$ from $t{=}0$ to $t{=}1$. Throughout, we write $\mu_\theta$ for the mean that the velocity field induces on one sampling step; it is the quantity our RL and distillation objectives act on.

\subsection{From ODE to SDE: enabling on-policy rollouts}
\label{sec:method:sde}

Deterministic ODE sampling maps a fixed noise sample to a single output and therefore offers no exploration, which reinforcement learning requires. We follow the practice of converting part of the sampling trajectory into a stochastic differential equation (SDE)~\citep{wang2026flowtts, liu2026flow}: for a subset of the integration steps we add Gaussian noise around the deterministic mean, so that one step produces a sampled next state $x'$, a deterministic mean $\mu_\theta(x_t, t)$, and the log-probability of the transition,
\begin{equation}
x' = \mu_\theta(x_t, t) + \sigma_t\,\epsilon,\qquad \epsilon \sim \mathcal{N}(0, I),
\qquad
\log \pi_\theta(x' \mid x_t) = -\frac{\lVert x' - \mu_\theta(x_t, t)\rVert^2}{2\sigma_t^2} + \text{const}.
\end{equation}
Sampling the same prompt several times now yields distinct rollouts, and each rollout records, at every SDE step, the state, condition, sampled output, and transition log-probability. The mean $\mu_\theta$ carries gradients and is the shared optimization target of both routes described below: the group-relative policy gradient and the distillation loss.

\subsection{Single-objective teachers via GRPO}
\label{sec:method:grpo}

We first train one expert per objective (IF, WER, and MOS) with group relative policy optimization~\citep{shao2024deepseekmath}. The three rewards are deliberately simple. The IF reward is a binary consistency verdict from a judge that decides whether an utterance matches its instruction, $r_{\mathrm{IF}}\in\{0,1\}$ (a $3$-vote majority); the WER reward is $r_{\mathrm{WER}}=1-\min(\mathrm{CER},1)$ from an ASR model; and the MOS reward is a no-reference quality score $r_{\mathrm{MOS}}\in[1,5]$. For each prompt we draw a group of $G$ on-policy rollouts, score each with the corresponding reward $r$, and standardize the scores within the group to obtain the advantage
\begin{equation}
A_i = \frac{r_i - \mathrm{mean}(\{r_j\}_{j=1}^{G})}{\mathrm{std}(\{r_j\}_{j=1}^{G})},
\end{equation}
so that only the relative quality within a group drives the update. The policy is then trained with the group-relative policy-gradient objective together with a KL regularizer to a frozen reference that prevents the base speech quality from drifting,
\begin{equation}
\mathcal{L}_{\mathrm{GRPO}} = -\,\mathbb{E}\big[A \,\log \pi_\theta\big] + \beta\, \mathrm{KL}\!\left(\pi_\theta \,\Vert\, \pi_{\mathrm{ref}}\right).
\end{equation}
This yields three single-objective experts, each strong on its own axis but weak on the others. Fusing them under a single scalar reward, however, requires collapsing the per-objective advantages into one number; the direction along which the objectives disagree is then lost, and optimization drifts toward the objective that is easiest to improve. This is the seesaw effect that motivates the distillation-based fusion below.

\subsection{Multi-objective fusion via on-policy distillation}
\label{sec:method:opd}

Rather than fusing the objectives through a scalar reward, we fuse the three experts directly in the velocity-field space with on-policy distillation~\citep{fang2026flow, li2026diffusionopd}, as illustrated in Figure~\ref{fig:overview}. The student generates its own on-policy rollouts, and each teacher is evaluated at the same trajectory points that the student visits, so the supervision is delivered exactly where the student's own sampling takes it.

Because the student and every teacher share the same flow-matching transition kernel, the two single-step transition distributions are Gaussians with the same covariance, and their per-step Kullback--Leibler (KL) divergence collapses in closed form to a squared distance between their means,
\begin{equation}
\mathrm{KL}\!\left(\pi_{\mathrm{student}} \,\Vert\, \pi_{\mathrm{teacher}}\right)
= \frac{\lVert \mu_{\mathrm{student}} - \mu_{\mathrm{teacher}} \rVert^2}{2\,\sigma_t^2\,\Delta t}.
\end{equation}
The distillation signal is therefore a fully differentiable, zero-variance regression on the velocity-field mean, and it needs neither advantages nor a policy-gradient estimator.

The key question is how to combine several teachers. We route each rollout to a single main teacher by the domain of its training data: rollouts drawn from the instruction-following data are distilled against the IF teacher, and rollouts from the WER data against the WER teacher, with the overall mixing set by the data ratio. The reason to route the primary objectives to \emph{disjoint} rollouts rather than have every rollout follow both teachers at once is the same conflict identified for the WER reward in Section~\ref{sec:exp:reward}: the WER data cannot contain dialect instructions, because a WER teacher scored by a Mandarin ASR would then push the model to abandon the requested dialect to lower its error rate, directly fighting instruction following. Keeping the IF and WER signals on separate rollouts prevents this tug-of-war on any single utterance. Quality, by contrast, has no domain of its own; it is a property every utterance should have. We therefore keep the MOS teacher out of the routing and add it as a light auxiliary term applied to all rollouts. The student minimizes
\begin{equation}
\mathcal{L}_{\mathrm{OPD}} =
\big\lVert \mu_{\mathrm{student}} - \mathrm{sg}(\mu_{d(i)}) \big\rVert^2
\;+\; \lambda_{\mathrm{MOS}}\,
\big\lVert \mu_{\mathrm{student}} - \mathrm{sg}(\mu_{\mathrm{MOS}}) \big\rVert^2,
\label{eq:opd}
\end{equation}
where $d(i)\in\{\mathrm{IF},\mathrm{WER}\}$ is the main teacher selected by the data domain of rollout $i$, $\mathrm{sg}(\cdot)$ is the stop-gradient operator that freezes the teachers as fixed targets, and $\lambda_{\mathrm{MOS}}$ is a small weight that makes quality a guardrail rather than a competing objective.

\subsection{Conflict-aware velocity fusion}
\label{sec:method:conflict}

Adding the two targets in Eq.~\ref{eq:opd} treats the primary target and the quality target as two independent regressions whose gradients are simply summed, which is benign only when the two point in compatible directions. But quality and instruction following often disagree on a single utterance: naturalness pushes toward a smooth, evenly-paced delivery, whereas following an expressive instruction pushes toward crying, trembling, or dialectal speech that a quality predictor reads as degradation. On the frames where the two disagree, the summed gradient partially cancels, and raising $\lambda_{\mathrm{MOS}}$ to buy quality drags the primary objective down, exactly the seesaw we set out to avoid. We quantify how often this happens in Section~\ref{sec:exp:conflict}.

We therefore fuse the two targets geometrically rather than additively. Writing the primary and quality update directions at a frame as
\begin{equation}
d_P = \mu_{d(i)} - \mu_{\mathrm{student}}, \qquad
d_U = \mu_{\mathrm{MOS}} - \mu_{\mathrm{student}},
\end{equation}
we remove from $d_U$ only the component that directly opposes $d_P$, keeping the aligned and orthogonal part,
\begin{equation}
d_U^{\mathrm{keep}} = d_U - \frac{\min\!\left(\langle d_U, d_P\rangle,\, 0\right)}{\lVert d_P\rVert^2}\, d_P,
\qquad
\mathcal{L}_{\mathrm{OPD}}^{\mathrm{conflict}} =
\big\lVert \mu_{\mathrm{student}} - \mathrm{sg}\!\left(\mu_{\mathrm{student}} + d_P + \lambda_{\mathrm{MOS}}\, d_U^{\mathrm{keep}}\right) \big\rVert^2 .
\end{equation}
When the two directions form an obtuse angle ($\langle d_U, d_P\rangle < 0$) the opposing component is projected out, so $d_U^{\mathrm{keep}} \perp d_P$ and quality can no longer erode the primary objective; when they are aligned ($\langle d_U, d_P\rangle \geq 0$) the quality push is kept in full. Quality is thus only ever allowed to make orthogonal adjustments: it may refine the timbre or smoothness of a frame, but it can never flatten out an already-expressed sob or dialect. This borrows the idea of gradient surgery~\citep{yu2020gradient}, but applies it to the per-frame velocity-field distillation targets of the flow model rather than to parameter gradients. It is an asymmetric, output-space geometric operation that a scalar-reward policy gradient cannot express, and that becomes available precisely because the distillation signal is a differentiable regression on $\mu$. Because the target is stop-gradient, the loss reduces to a single mean-squared error and $\lambda_{\mathrm{MOS}}$ scales the surviving quality push linearly; \methodname{} uses this conflict-aware form as its fusion loss.

\section{Experiments}
\label{sec:exp}

\subsection{Setup}
\label{sec:exp:setup}

\textbf{Data and base model.} All models start from a single supervised DiTAR checkpoint (the base) and are post-trained with RL and OPD on $30{,}000$ instruction-following and $30{,}000$ intelligibility (WER) utterances, a $1{:}1$ ratio. Every instruction is in the ZH-DSD style, matching the base model, which has only seen ZH-DSD-style instruction data. We therefore evaluate primarily on the DSD subset of InstructTTSEval~\citep{huang2025instructttseval} (zh\_DSD, $1000$ utterances), and also report the full six InstructTTSEval subsets to probe generalization.

\textbf{Metrics.} Each axis is optimized against a single chosen reward, but we report every axis with \emph{both} its reward-side metric and a held-out metric that never shapes training, so that reward hacking would surface as a divergence between the two. Instruction following (IF), the primary axis, is the consistency rate of a judge that decides whether an utterance matches its instruction: a distilled Qwen2.5-Omni proxy (IF$_{\mathrm{qwen}}$) for analysis and checkpoint selection, and Gemini as the held-out judge. Intelligibility is the character error rate (CER, $\downarrow$) from Paraformer-zh under the seed-tts-eval protocol, held out against Qwen3-ASR (CER$_{\mathrm{q3}}$) and reported de-dialect. Quality is DNSMOS P.835 OVRL ($\uparrow$), held out against UTMOS.

\textbf{Training.} Teachers are trained with GRPO and fused with OPD on $4$ nodes $\times\,8$ GPUs for $350$ steps. GRPO draws $G{=}32$ rollouts per prompt with a KL coefficient $\beta{=}0.01$; OPD uses $16$ rollouts, needs no reward server, and sets the quality guardrail weight to $\lambda_{\mathrm{MOS}}{=}0.3$. Both use a learning rate of $2\!\times\!10^{-6}$. Full hyperparameters are in Appendix~\ref{app:hparams}.

\subsection{Reward analysis}
\label{sec:exp:reward}

OPD fuses three single-objective teachers, so the quality of the fusion rests on having a strong teacher for each axis, which in turn depends on a sound reward. We therefore carefully analyze the reward choice along each of the three axes before training the teachers.

\begin{figure}[t]
\centering
\begin{minipage}{0.48\textwidth}
  \centering
  \includegraphics[width=\linewidth]{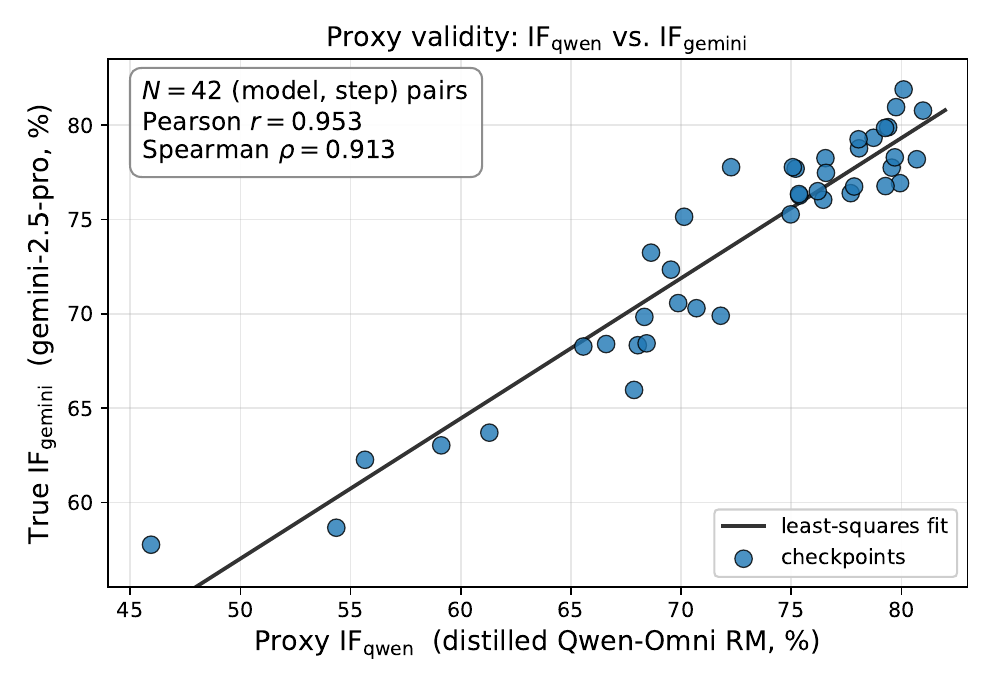}
  \caption{Distilled Qwen-Omni IF proxy vs.\ the Gemini judge, one point per
  (model, step) checkpoint scored on zh\_DSD.
  $x$ = proxy IF$_{\mathrm{qwen}}$, $y$ = Gemini IF.}
  \label{fig:proxy}
\end{minipage}\hfill
\begin{minipage}{0.48\textwidth}
  \centering
  \includegraphics[width=\linewidth]{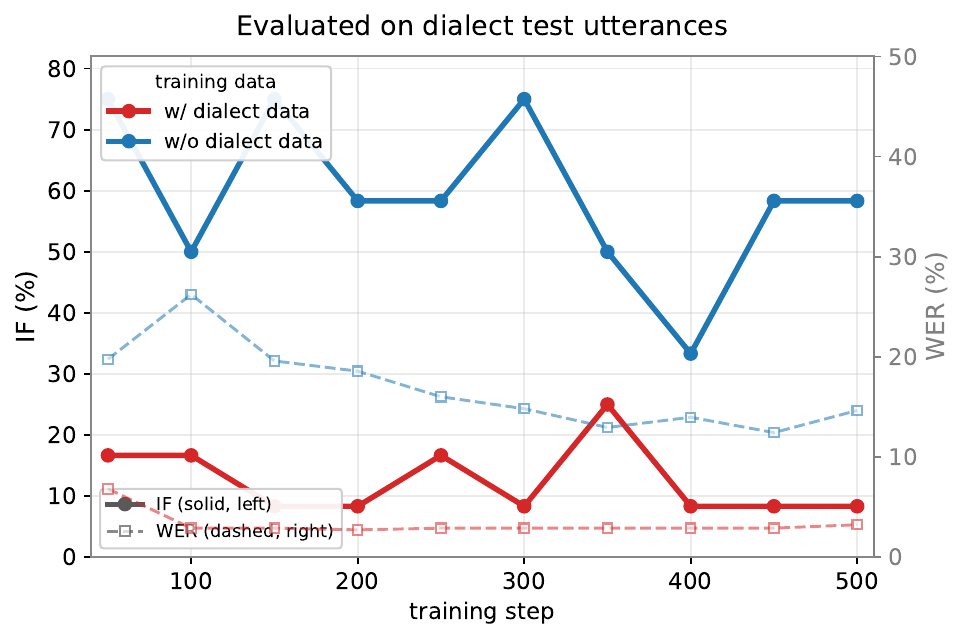}
  \caption{WER teacher's performance on the dialect (Cantonese) test subset
  across training, with vs.\ without dialect data. Solid lines are IF judged by
  Gemini on the left axis; dashed lines are WER on the right axis.}
  \label{fig:dialect}
\end{minipage}
\end{figure}

\begin{table}[t]
\centering
\begin{minipage}{0.44\textwidth}
  \centering
  \caption{ASR back-end for the WER reward: Corpus CER (\%) on the full zh\_DSD
  set (with dialect). Base = our supervised model; GT = human recordings.}
  \label{tab:asr}
  \begin{tabular}{lcc}
  \toprule
  Speech & Paraformer-zh & Qwen3-ASR \\
  \midrule
  Base & 3.51 & 3.17 \\
  GT   & 9.84 & 8.12 \\
  \bottomrule
  \end{tabular}
\end{minipage}\hfill
\begin{minipage}{0.52\textwidth}
  \centering
  \caption{Language bias of MOS predictors on matched clean human recordings
  (seed-tts-eval).}
  \label{tab:mosbias}
  \setlength{\tabcolsep}{4pt}
  \begin{tabular}{lccc}
  \toprule
  MOS predictor & zh & en & $|$gap$|$ / range \\
  \midrule
  UTMOS               & 2.774 & 3.491 & \textbf{17.9\%} \\
  DNSMOS OVRL         & 3.183 & 3.071 & 2.8\% \\
  DNSMOS SIG          & 3.587 & 3.455 & 3.3\% \\
  DNSMOS BAK          & 3.854 & 3.820 & 0.8\% \\
  \bottomrule
  \end{tabular}
\end{minipage}
\end{table}

\textbf{Instruction following: a distilled proxy for the Gemini judge.} The standard way to score IF for Instruct-TTS is Gemini-as-a-judge~\citep{ yang2024instructtts, chen2026mint, li2026anyaudio}, which is reliable but far too expensive to query for the millions of rollouts an RL run consumes. We therefore distill a Qwen2.5-Omni-7B~\citep{xu2025qwen25omni} reward model on Gemini judgments and ask whether it can serve as the IF reward in its place. Figure~\ref{fig:proxy} scores $42$ checkpoints spanning a wide IF range with both judges and plots one against the other: the points fall close to a straight line, and the two judgments are strongly correlated (Pearson $r{=}0.95$, Spearman $\rho{=}0.91$). The distilled model thus ranks systems and checkpoints consistently with Gemini, so we use it as a low-cost reward model in place of Gemini throughout training and development.

\textbf{Intelligibility: the WER teacher's training data must exclude dialect.} A WER reward scored by a Mandarin ASR conflicts with instruction following whenever the instruction asks for a dialect, and Figure~\ref{fig:dialect} makes this concrete. Trained \emph{with} dialect data, the WER teacher drives its Mandarin-ASR WER down sharply, yet its IF stays near $10\%$: the model has simply stopped speaking the requested dialect and reads plain Mandarin that the ASR transcribes happily, so the WER gain is spurious, bought by disobeying the instruction. Trained \emph{without} dialect data, it keeps the dialect and its IF is three to seven times higher, even though the Mandarin ASR reports a much higher WER on those correct utterances. We therefore remove dialect from the WER teacher's training data and report CER de-dialect throughout. We also tried a stronger ASR back-end for expressive and dialectal speech, but Table~\ref{tab:asr} shows Paraformer-zh~\citep{gao2022paraformer} and Qwen3-ASR~\citep{shi2026qwen3} nearly indistinguishable ($0.1$--$1$\,pp CER) on both generated and human speech, so we keep Paraformer-zh.

\textbf{Quality: UTMOS is language-biased, DNSMOS is not.} A quality reward is only safe if it scores speech, not language. Table~\ref{tab:mosbias} scores matched clean human recordings in Chinese and English, whose true quality is comparable, and reads off the gap each predictor introduces between the two. UTMOS~\citep{saeki2022utmos} rates English $0.72$ points higher than Chinese, $18\%$ of its range; this bias reflects its almost-entirely-English training data rather than the audio, and as a reward it would push the model toward a foreign-accented reading of Chinese. DNSMOS P.835~\citep{reddy2022dnsmos} shows almost no such gap ($\le 3\%$ across all sub-scores), so we adopt DNSMOS OVRL as both the quality reward and the reported metric.

\subsection{Main Results}
\label{sec:exp:main}

\begin{table}[t]
\centering
\caption{Main comparison on the zh\_DSD subset of InstructTTSEval. Avg is the
mean of IF, CER, and DNSMOS after min--max normalizing each to $[0,1]$ with CER
inverted. Per column, best in \textbf{bold}, second \underline{underlined}.}
\label{tab:main}
\begin{tabular}{lcccccc}
\toprule
System & IF $\uparrow$ & CER $\downarrow$ & CER$_{\mathrm{q3}}$ $\downarrow$ & DNSMOS $\uparrow$ & UTMOS $\uparrow$ & Avg $\uparrow$ \\
\midrule
Ground-truth (human)          & 78.5 & 8.70 & 7.69 & 3.00 & 1.73 & 0.36 \\
Base (supervised)             & 72.2 & 3.18 & 3.08 & 2.81 & 1.95 & 0.48 \\
\midrule
Teacher: IF                   & \textbf{82.1} & 2.53 & 2.40 & 2.98 & 1.99 & 0.70 \\
Teacher: WER                  & 71.0 & \textbf{1.94} & \textbf{2.00} & 3.25 & \underline{2.65} & 0.70 \\
Teacher: MOS (quality)        & 55.0 & 2.60 & 2.51 & \textbf{3.70} & \textbf{3.62} & 0.63 \\
\midrule
% GRPO multi-reward (weighted)  & 77.0 & 2.34 & 3.25 \\
% GRPO multi-reward (routing)   & 72.6 & 2.27 & 3.29 \\
GRPO multi-reward             & 77.0 & 2.34 & 2.27 & 3.25 & 2.48 & 0.75 \\
\methodname{}                 & \underline{81.1} & \underline{2.25} & \underline{2.21} & 3.28 & 2.57 & \textbf{0.82} \\
\quad w/o conflict-aware fusion & 79.4 & 2.29 & 2.23 & \underline{3.31} & \underline{2.65} & \underline{0.80} \\
\bottomrule
\end{tabular}
\end{table}

With the three rewards fixed, we train one GRPO teacher per objective and then fuse them. As Table~\ref{tab:main} shows, each teacher is strong on its own axis and visibly weak on the others: the IF teacher reaches $82.1$ IF but only middling CER and quality, the WER teacher drives CER down to $1.94$ at the cost of IF, and the MOS teacher tops DNSMOS while its IF collapses to $55$.

We compare three fusion strategies under matched teachers and data. The scalar-reward baseline (\emph{GRPO multi-reward}) still seesaws even with group-decoupled normalization~\citep{liu2026gdpo}, improving the easier axes of quality and intelligibility while leaving the primary objective, IF, the lowest among all fused systems. On-policy distillation fares much better: our full method attains the highest IF of any fused system, close to the IF teacher's, together with the best CER while keeping DNSMOS strong. Removing conflict-aware fusion (\emph{w/o conflict-aware fusion}) drops IF, since the quality target then conflicts with instruction following. Here Avg is the mean of the three optimized objectives after min--max normalizing each to $[0,1]$ with CER inverted, and the held-out checks CER$_{\mathrm{q3}}$ and UTMOS are excluded from it.

To confirm these gains are genuine rather than reward overfitting, Table~\ref{tab:main} also reports two metrics that never shape training: CER from a different ASR back-end (CER$_{\mathrm{q3}}$) and UTMOS. Both preserve the same ordering across systems as their reward-side counterparts, confirming that the gains are real. UTMOS values are uniformly low because it is biased against Chinese, so we read it only for the ranking it induces, which is exactly why DNSMOS is our quality reward and headline metric.

\subsection{Results on InstructTTSEval}
\label{sec:exp:instructttseval}

Although our model has only ever seen instructions in the ZH-DSD style, we want to know whether its instruction-following ability generalizes to unseen instruction forms. We therefore evaluate on all six subsets of InstructTTSEval, \{ZH, EN\} $\times$ \{APS, DSD, RP\}, scoring every system with the standard Gemini judge for a fair comparison; the three ZH subsets use the original instructions and the three EN subsets use their Chinese translations. Table~\ref{tab:instructttseval} reports the results against both open and closed baselines. Post-training improves over the base on every subset, with the largest gains on the in-distribution DSD style and smaller but consistent gains on the harder APS and RP styles and on the out-of-distribution English side; among open systems our model is strongest or second on the Chinese subsets and competitive on English, while a gap to the closed Gemini systems remains on the RP style and on English. The improvements thus hold across languages and instruction styles well beyond the training distribution, showing that the fused model generalizes rather than overfitting the ZH-DSD form it was trained on.

%\begin{table}[t]
%\centering
%\caption{Instruction following on InstructTTSEval (Gemini judge, \%, $\uparrow$).
%CV = CustomVoice, VD = VoiceDesign. Best among open systems in bold, second
%underlined.}
%\label{tab:instructttseval}
%\begin{tabular}{lc}
%\toprule
%Model & IF (\%) $\uparrow$ \\
%\midrule
%Gemini-flash~\citep{comanici2025gemini}    & 90.9 \\
%Gemini-pro~\citep{comanici2025gemini}       & 90.1 \\
%GPT-4o-mini-tts~\citep{hurst2024gpt}        & 52.3 \\
%\midrule
%VoxInstruct~\citep{zhou2024voxinstruct}     & 52.3 \\
%VoiceSculptor~\citep{hu2026voicesculptor}   & 64.7 \\
%Mimo-Audio-7B-Instruct~\citep{xiaomi2025mimo} & 74.3 \\
%Qwen3TTS-25Hz-1.7B-CV~\citep{hu2026qwen3}   & 75.0 \\
%Qwen3TTS-12Hz-1.7B-CV~\citep{hu2026qwen3}   & 77.8 \\
%Qwen3TTS-12Hz-1.7B-VD~\citep{hu2026qwen3}   & 81.1 \\
%Base (Ours, before RL)              & 76.9 \\
%IF teacher (Ours)                   & \textbf{83.9} \\
%\methodname{} (Ours)                & \underline{82.1} \\
%\bottomrule
%\end{tabular}
%\end{table}

\begin{table}[t]
\centering
\caption{Instruction following on InstructTTSEval (Gemini judge, \%, $\uparrow$),
across all six subsets: \{ZH, EN\} $\times$ \{APS, DSD, RP\}. CV = CustomVoice,
VD = VoiceDesign. Best among open systems in bold, second underlined.}
\label{tab:instructttseval}
\setlength\tabcolsep{4pt}
\resizebox{0.9\textwidth}{!}{%
\begin{tabular}{lcccccc}
\toprule
\textbf{Model} & \multicolumn{3}{c}{\textbf{InstructTTSEval-ZH}} & \multicolumn{3}{c}{\textbf{InstructTTSEval-EN}} \\
\cmidrule(lr){2-4}\cmidrule(lr){5-7}
 & APS ($\uparrow$) & DSD ($\uparrow$) & RP ($\uparrow$) & APS ($\uparrow$) & DSD ($\uparrow$) & RP ($\uparrow$) \\
\midrule
Gemini-flash~\citep{comanici2025gemini}    & 88.2 & 90.9 & 77.3 & 92.3 & 93.8 & 80.1 \\
Gemini-pro~\citep{comanici2025gemini}       & 89.0 & 90.1 & 75.5 & 87.6 & 86.0 & 67.2 \\
GPT-4o-mini-tts~\citep{hurst2024gpt}        & 54.9 & 52.3 & 46.0 & 76.4 & 74.3 & 54.8 \\
\midrule
VoxInstruct~\citep{zhou2024voxinstruct}     & 47.5 & 52.3 & 42.6 & 54.9 & 57.0 & 39.3 \\
VoiceSculptor~\citep{hu2026voicesculptor}   & 75.7 & 64.7 & 61.5 & -- & -- & -- \\
Mimo-Audio-7B-Instruct~\citep{xiaomi2025mimo} & 75.7 & 74.3 & 61.5 & 80.6 & 77.6 & 59.5 \\
Qwen3TTS-25Hz-1.7B-CV~\citep{hu2026qwen3}   & 83.1 & 75.0 & \underline{63.0} & 79.0 & \textbf{82.8} & \textbf{69.3} \\
Qwen3TTS-12Hz-1.7B-CV~\citep{hu2026qwen3}   & 83.0 & 77.8 & 61.2 & 77.3 & 77.1 & 63.7 \\
Qwen3TTS-12Hz-1.7B-VD~\citep{hu2026qwen3}   & 85.2 & 81.1 & \textbf{65.1} & \textbf{82.9} & \underline{82.4} & \underline{68.4} \\
Base (Ours, before RL)              & 86.0 & 76.9 & 54.4 & 79.2 & 69.5 & 46.4 \\
IF teacher (Ours)                   & \textbf{90.9} & \textbf{83.9} & 58.7 & \underline{81.3} & 75.6 & 48.8 \\
\methodname{} (Ours)                & \underline{90.6} & \underline{82.1} & 56.2 & 80.8 & 73.4 & 47.2 \\
\bottomrule
\end{tabular}%
}
\end{table}

\section{Analysis}
\label{sec:analysis}

\subsection{Per-frame conflict geometry}
\label{sec:exp:conflict}

\begin{figure}[t]
\centering
\includegraphics[width=0.9\linewidth]{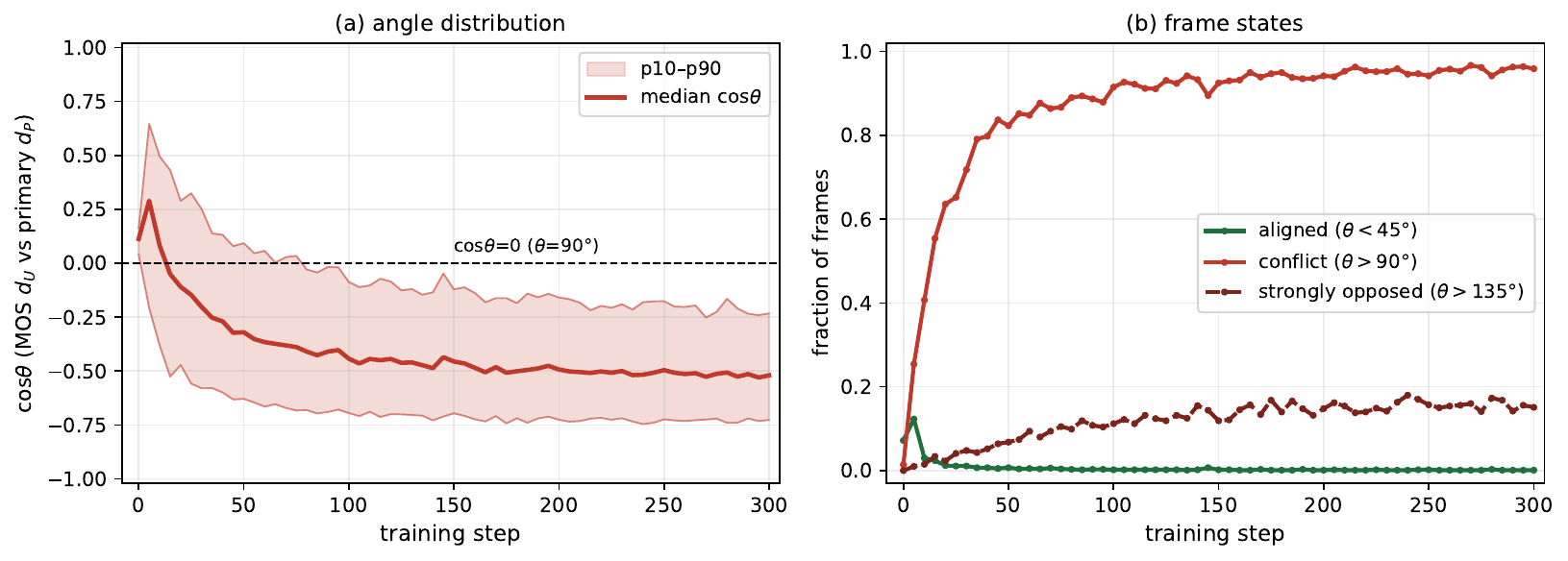}
\caption{Per-frame geometry between the quality direction $d_U$ and the primary
direction $d_P$ over training. \textbf{(a)} the cosine-similarity distribution,
shown as the median with a 10 to 90\% band; \textbf{(b)} the fractions of aligned
($\theta<45^\circ$), conflicting ($\theta>90^\circ$), and strongly-opposed
($\theta>135^\circ$) frames.}
\label{fig:conflict-angle}
\end{figure}

Conflict-aware fusion projects out the component of the quality push $d_U$ that opposes the primary push $d_P$, so whether it matters depends on how often the two directions actually conflict. We measure this directly. Figure~\ref{fig:conflict-angle} tracks the per-frame angle $\theta$ between $d_U$ and $d_P$ over training: within about fifteen steps the median cosine similarity turns negative, aligned frames collapse to essentially zero, and conflicting frames climb to nearly all of them, with strongly-opposed frames still growing thereafter. Once the student becomes expressive, quality opposes the primary objective on almost every frame. This near-total conflict is exactly why the fusion rule matters: adding the quality push without projection would carry the opposing component almost everywhere and pull instruction following down, whereas the projection removes it frame by frame and keeps instruction following near the teacher's ceiling.

\subsection{Trade-offs Analysis}
\label{sec:exp:tradeoff}

The conflict above is about gradients; we now ask whether it surfaces in the end metrics, and find that the answer depends sharply on which pair of objectives we look at. We sweep two knobs and read off the frontiers in Figure~\ref{fig:tradeoff}, with full numbers in Appendix~\ref{app:sweeps}. Sweeping the instruction:WER data ratio traces out an essentially flat IF--WER frontier: adding WER data drives CER toward its floor without costing instruction following, so the two improve together with no meaningful trade-off, as expected when they are routed to disjoint rollouts. A further ablation in Appendix~\ref{sec:exp:ablation} suggests that this routing, rather than the dense velocity supervision alone, is the main factor preserving instruction following. Sweeping the quality weight $\lambda_{\mathrm{MOS}}$ tells the opposite story. Without conflict-aware fusion, instruction following falls steadily as the quality pressure rises, the metric-level shadow of the near-total per-frame conflict; conflict-aware velocity fusion flattens this trade-off, keeping instruction following far higher at matched weights because the projection stops the quality push from eroding it frame by frame.

\begin{figure}[t]
\centering
\begin{minipage}{0.9\textwidth}
\begin{minipage}{0.49\textwidth}
  \centering
  \includegraphics[width=\linewidth]{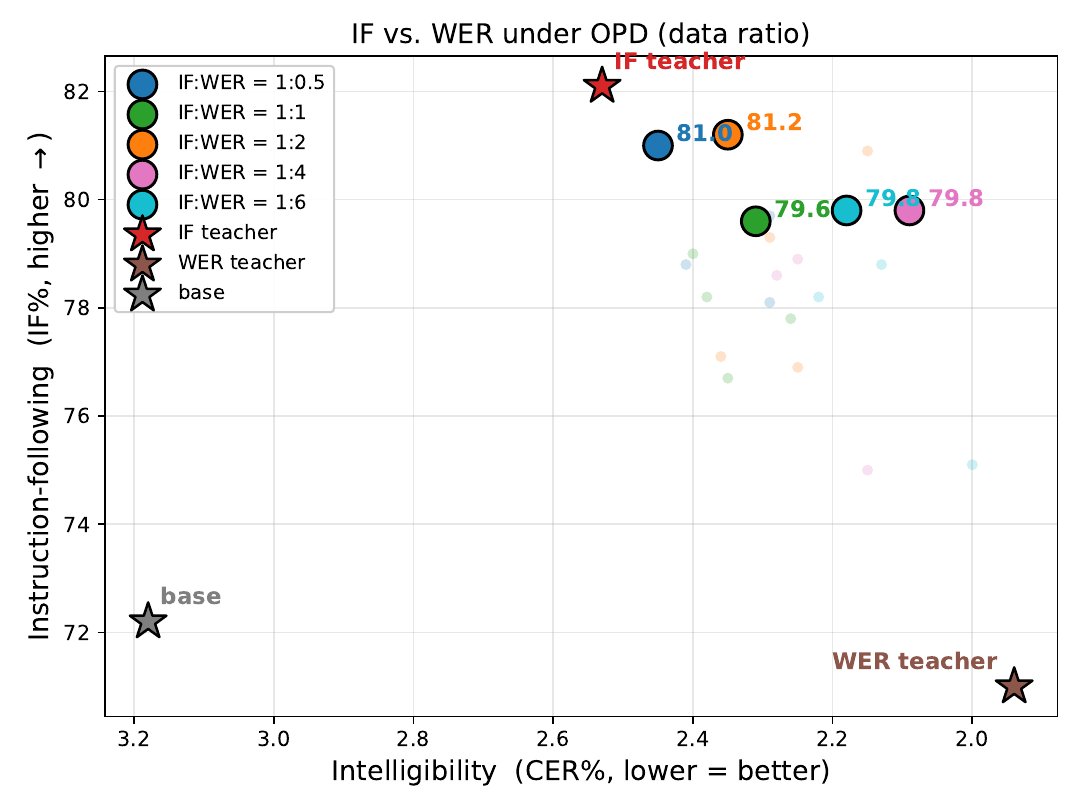}
\end{minipage}\hfill
\begin{minipage}{0.49\textwidth}
  \centering
  \includegraphics[width=\linewidth]{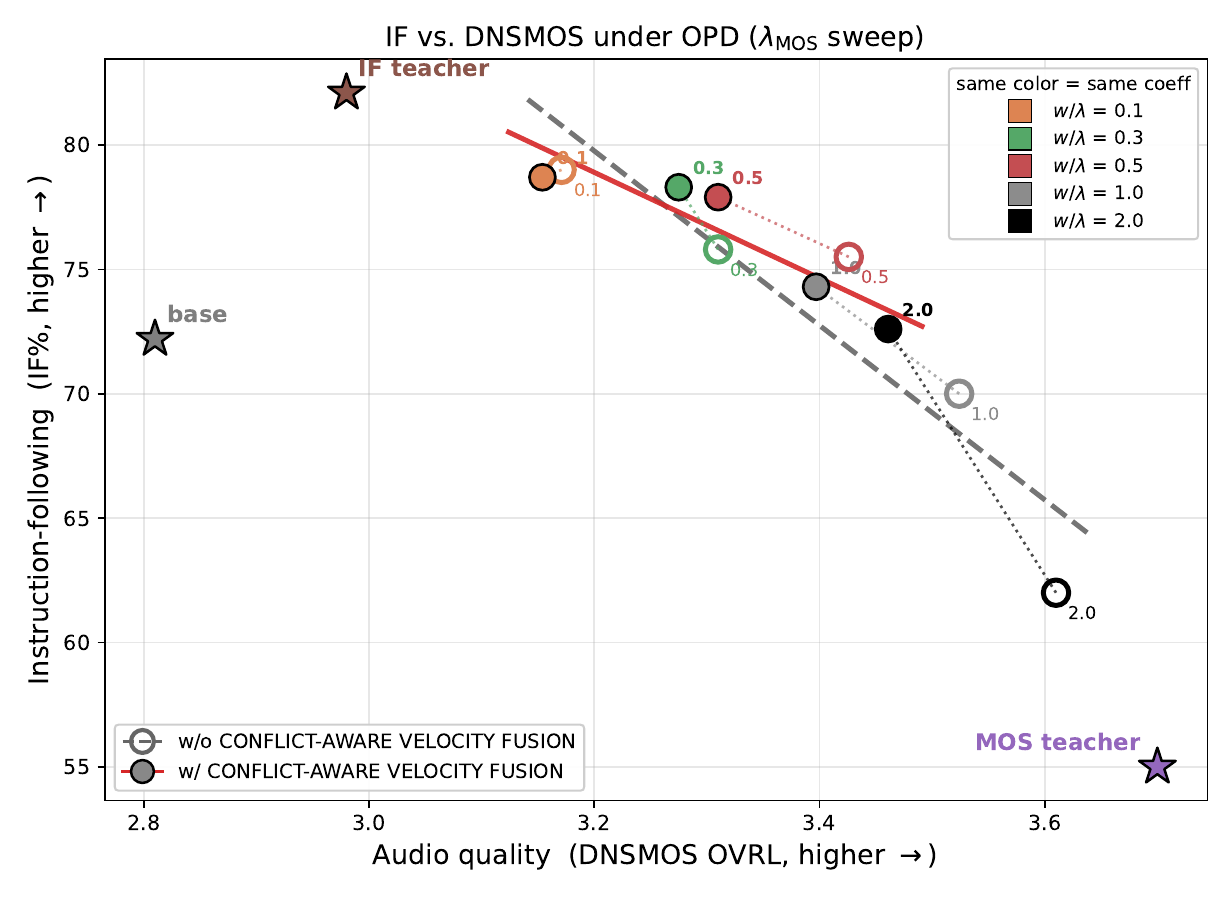}
\end{minipage}
\end{minipage}
\caption{Trade-offs under OPD. \textbf{Left:} IF against WER while sweeping the
instruction:WER data ratio; large dots mark the best-IF checkpoint per ratio,
faint dots all checkpoints. \textbf{Right:} IF against DNSMOS while sweeping the
quality weight $\lambda_{\mathrm{MOS}}$, with (filled) and without (hollow)
conflict-aware velocity fusion; same color shares a $\lambda_{\mathrm{MOS}}$,
stars mark the teachers and base.}
\label{fig:tradeoff}
\end{figure}

\section{Conclusion}
\label{sec:conclusion}

We studied how to post-train an instruction-following TTS model on its three competing objectives without the seesaw a single scalar reward induces, and two lessons emerged. First, the reward for each axis matters as much as the fusion algorithm: a distilled Qwen-Omni proxy tracks the costly Gemini judge, the WER reward must be dialect-free, and DNSMOS replaces the language-biased UTMOS. Second, fusing single-objective teachers through on-policy distillation improves all three axes together where a scalar reward cannot: domain routing lets instruction following and intelligibility rise together, while quality, which conflicts with instruction following, is fused geometrically rather than additively. This per-frame geometric fusion is available only because distillation in continuous flow-matching space is a differentiable regression on the velocity field, and we hope it offers a practical recipe for multi-objective alignment of expressive speech generators.

%\subsubsection*{Author Contributions}
%If you'd like to, you may include  a section for author contributions as is done in many journals. This is optional and at the discretion of the authors.
%
%\subsubsection*{Acknowledgments}
%Use unnumbered third level headings for the acknowledgments. All acknowledgments, including those to funding agencies, go at the end of the paper.

\subsection*{AI use statement}
We used generative AI tools to polish the writing of this paper, improving its grammar and readability.

\bibliography{iclr2027_conference}
\bibliographystyle{iclr2027_conference}

\clearpage
\appendix
\section{Hyperparameters}
\label{app:hparams}

Table~\ref{tab:hparams} lists the training hyperparameters for the GRPO teachers and for OPD fusion. All runs start from the same supervised base checkpoint and train on $4$ nodes $\times\,8$ GPUs. The flow-matching decoder uses a patch size of $5$ latents per frame with a $10$-step ODE, of which a contiguous $3$-step window is converted to an SDE for on-policy rollouts, under classifier-free guidance $\alpha{=}2.0$.

\begin{table}[h]
\centering
\caption{Training hyperparameters. GRPO trains one single-objective teacher each
(IF, WER, MOS); OPD fuses them.}
\label{tab:hparams}
\resizebox{0.6\linewidth}{!}{%
\begin{tabular}{lcc}
\toprule
 & GRPO (per teacher) & OPD fusion \\
\midrule
Rollouts per prompt $G$        & 32       & 16 \\
KL coefficient $\beta$         & 0.01     & 0 \\
Learning rate                  & $2\times10^{-6}$ & $2\times10^{-6}$ \\
Optimizer                      & Adam ($0.9,0.95$) & Adam ($0.9,0.95$) \\
Micro-batch / grad-accum       & 4 / 4    & 4 / 4 \\
Checkpoint interval (steps)    & 50       & 50 \\
Reward server required         & yes      & no \\
Fusion mode                    & ---      & domain routing \\
MOS guardrail $\lambda_{\mathrm{MOS}}$ & --- & 0.3 \\
\midrule
\multicolumn{3}{l}{\emph{Shared flow-matching / SDE settings}} \\
Latents per frame (patch size) & \multicolumn{2}{c}{5} \\
ODE steps / SDE window         & \multicolumn{2}{c}{10 / 3} \\
CFG strength $\alpha$          & \multicolumn{2}{c}{2.0} \\
\bottomrule
\end{tabular}%
}
\end{table}

\section{Trade-off sweeps: full numbers}
\label{app:sweeps}

This appendix gives the full numbers behind the two trade-off plots of
Figure~\ref{fig:tradeoff}, with the single-objective teachers as the two extremes
of each sweep. Table~\ref{tab:sweep_wer} lists the instruction:WER data-ratio
sweep, reporting the best-IF checkpoint per ratio. Table~\ref{tab:sweep_mos} lists
the quality-weight $\lambda_{\mathrm{MOS}}$ sweep, reporting both fusion rules,
without and with conflict-aware velocity fusion, at a matched checkpoint
(step~200) so that only the fusion rule differs between the two halves. All runs
use the final three teachers and are evaluated on zh\_DSD ($n{=}1000$), with IF
from the Qwen2.5-Omni RM (\%, $\uparrow$), CER from Paraformer-zh (de-dialect,
\%, $\downarrow$), and DNSMOS P.835 OVRL ($\uparrow$).

\begin{table}[h]
\centering
\caption{Instruction:WER data-ratio sweep ($\lambda_{\mathrm{MOS}}{=}0.1$ fixed),
best-IF checkpoint per ratio.}
\label{tab:sweep_wer}
\begin{tabular}{lccc}
\toprule
IF:WER & IF $\uparrow$ & CER $\downarrow$ & DNSMOS $\uparrow$ \\
\midrule
IF teacher  & 82.1 & 2.53 & 2.98 \\
1:0.5 & 81.0 & 2.45 & 3.17 \\
1:1   & 79.6 & 2.31 & 3.18 \\
1:2   & 81.2 & 2.35 & 3.19 \\
1:4   & 79.8 & 2.09 & 3.23 \\
1:6   & 79.8 & 2.18 & 3.21 \\
WER teacher & 71.0 & 1.94 & 3.25 \\
\bottomrule
\end{tabular}
\end{table}

\begin{table}[h]
\centering
\caption{Quality-weight $\lambda_{\mathrm{MOS}}$ sweep (IF:WER${=}1{:}1$ fixed),
without vs.\ with conflict-aware velocity fusion at a matched checkpoint
(step~200). Everything but the fusion rule is held fixed, and teachers are the
two extremes of the sweep.}
\label{tab:sweep_mos}
\begin{tabular}{lccccccc}
\toprule
 & \multicolumn{3}{c}{w/o conflict-aware} & & \multicolumn{3}{c}{w/ conflict-aware} \\
\cmidrule(lr){2-4}\cmidrule(lr){6-8}
$\lambda_{\mathrm{MOS}}$ & IF $\uparrow$ & CER $\downarrow$ & DNSMOS $\uparrow$ & & IF $\uparrow$ & CER $\downarrow$ & DNSMOS $\uparrow$ \\
\midrule
IF teacher   & 82.1 & 2.53 & 2.98 & & 82.1 & 2.53 & 2.98 \\
0.1  & 79.0 & 2.40 & 3.17 & & 78.7 & 2.24 & 3.15 \\
0.3  & 75.8 & 2.33 & 3.31 & & 78.3 & 2.23 & 3.28 \\
0.5  & 75.5 & 2.28 & 3.43 & & 77.9 & 2.29 & 3.31 \\
1.0  & 70.0 & 2.32 & 3.52 & & 74.3 & 2.22 & 3.40 \\
2.0  & 62.0 & 2.40 & 3.61 & & 72.6 & 2.29 & 3.46 \\
MOS teacher  & 55.0 & 2.60 & 3.70 & & 55.0 & 2.60 & 3.70 \\
\bottomrule
\end{tabular}
\end{table}

\section{Ablation: is domain routing necessary?}
\label{sec:exp:ablation}

\begin{figure}[h]
\centering
\includegraphics[width=0.6\linewidth]{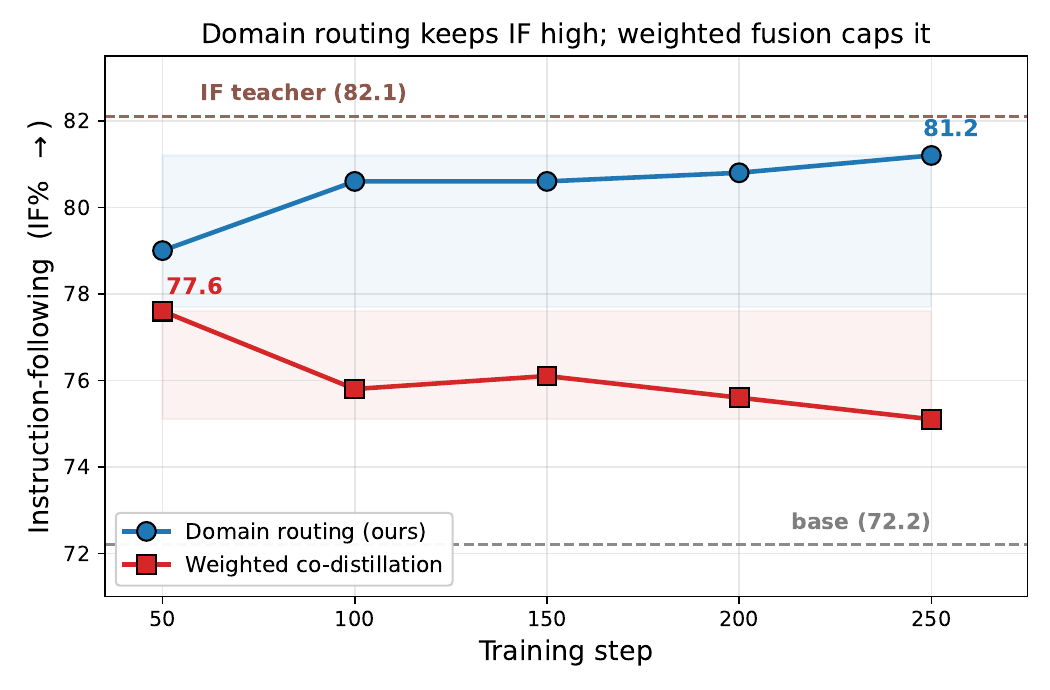}
\caption{Domain routing vs.\ weighted co-distillation: IF (Qwen-Omni RM) at
every checkpoint through convergence, under identical teachers and 1:1 data.}
\label{fig:ablation}
\end{figure}

The fusion in Section~\ref{sec:method:opd} routes each rollout to a single main teacher by its data domain. A natural alternative is to distill every rollout against all teachers at once with a weighted-sum loss, the velocity-field analog of scalar-reward fusion. We compare the two under identical teachers, data, and budget (Figure~\ref{fig:ablation}). Routing keeps IF high throughout training, climbing toward the IF teacher, whereas weighted co-distillation stays several points lower and drifts downward from the first checkpoints, so the two IF trajectories never overlap. Both keep CER and DNSMOS within target, so the gap is concentrated on IF, the primary objective. The mechanism is the same intra-sample conflict as in the scalar case: when a single utterance is pulled toward instruction following, intelligibility, and quality at once, the competing velocities average into a bland compromise that caps IF. Routing sidesteps this by sending the competing primary objectives to disjoint rollouts, which suggests that it is the routing, not merely the dense velocity supervision, that makes the fusion work.

\section{Human preference evaluation}
\label{app:human}

\begin{figure}[h]
\centering
\includegraphics[width=0.6\linewidth]{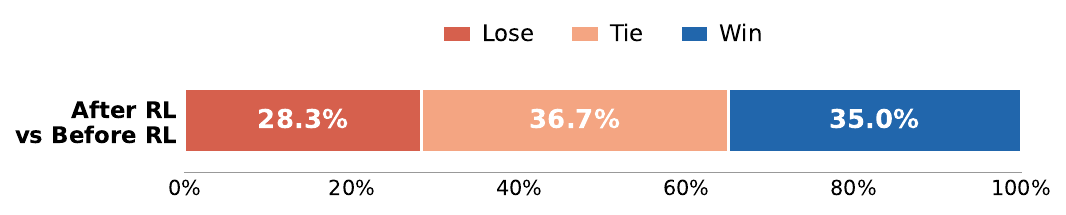}
\caption{Blind A/B human preference, after RL vs.\ before RL. Bars show the share
of comparisons the after-RL model loses, ties, and wins.}
\label{fig:human_pref}
\end{figure}

To check that the automatic-metric gains translate into perceived quality, we run
a blind A/B listening test comparing the final fused model after RL against the
supervised base checkpoint before RL. Raters hear the two utterances for each
prompt in randomized order and choose which one better follows the instruction
while sounding natural, or mark a tie. Aggregating over all trials, the after-RL
model wins $35.0\%$ of comparisons, loses $28.3\%$, and ties $36.7\%$
(Figure~\ref{fig:human_pref}), so it is preferred over the base in $6.7$ points
more comparisons than it loses, confirming that the objective-metric improvements
carry over to human judgement.

\end{document}